\documentclass[10pt,twocolumn,letterpaper]{article}

\usepackage{cvpr}
\usepackage{times}
\usepackage{epsfig}
\usepackage{graphicx}
\usepackage{amsmath}
\usepackage{pifont}
\usepackage{amssymb}
\usepackage{multirow}
\usepackage{sidecap}
\usepackage{subcaption}
\usepackage{algorithm}
\usepackage{algorithmic}

\newcommand{\myparagraph}[1]{\vspace{5pt}\noindent{\bf #1}}

\usepackage[breaklinks=true,bookmarks=false]{hyperref}

\cvprfinalcopy 

\def\cvprPaperID{3745} 

\begin{document}

\title{\texttt{f-VAEGAN-D2}: A Feature Generating Framework for Any-Shot Learning}

\author{
 Yongqin Xian$^{1}$ \hspace{4mm} Saurabh Sharma$^{1}$ \hspace{4mm} Bernt Schiele$^{1}$ \hspace{4mm} Zeynep Akata$^{1,2}$\vspace{4mm} \\ 
  \begin{tabular}{cc}
  $^{1}$Max Planck Institute for Informatics & $^{2}$Amsterdam Machine Learning Lab \\ Saarland Informatics Campus & University of Amsterdam 
 \end{tabular}
 }

\maketitle

\begin{abstract} 
When labeled training data is scarce, a promising data augmentation approach is to generate visual features of unknown classes using their attributes. To learn the class conditional distribution of CNN features, these models rely on pairs of image features and class attributes. Hence, they can not make use of the abundance of unlabeled data samples. In this paper, we tackle any-shot learning problems i.e. zero-shot and few-shot, in a unified feature generating framework that operates in both inductive and transductive learning settings. We develop a conditional generative model that combines the strength of VAE and GANs and in addition, via an unconditional discriminator, learns the marginal feature distribution of unlabeled images. We empirically show that our model learns highly discriminative CNN features for five datasets, i.e. CUB, SUN, AWA and ImageNet, and establish a new state-of-the-art in any-shot learning, i.e. inductive and transductive (generalized) zero- and few-shot learning settings. We also demonstrate that our learned features are interpretable: we visualize them by inverting them back to the pixel space and we explain them by generating textual arguments of why they are associated with a certain label.
\end{abstract}

\section{Introduction}
Learning with limited labels has been an important topic of research as it is unrealistic to collect sufficient amounts of labeled data for every object. Recently, generating visual features of previously unseen classes~\cite{XLSA18,BHJ17,Verma_2018_CVPR,FKRC18} has shown its potential to perform well on extremely imbalanced image collections. However, current feature generation approaches have still shortcomings. First, they rely on simple generative models which are not able to capture complex data distributions. Second, in many cases, they do not truly generalize to the under represented classes. Third, although classifiers trained on a combination of real and generated features obtain state-of-the-art results, generated features may not be easily interpretable.

\begin{figure}[t]
	\centering
        \includegraphics[width=\columnwidth]{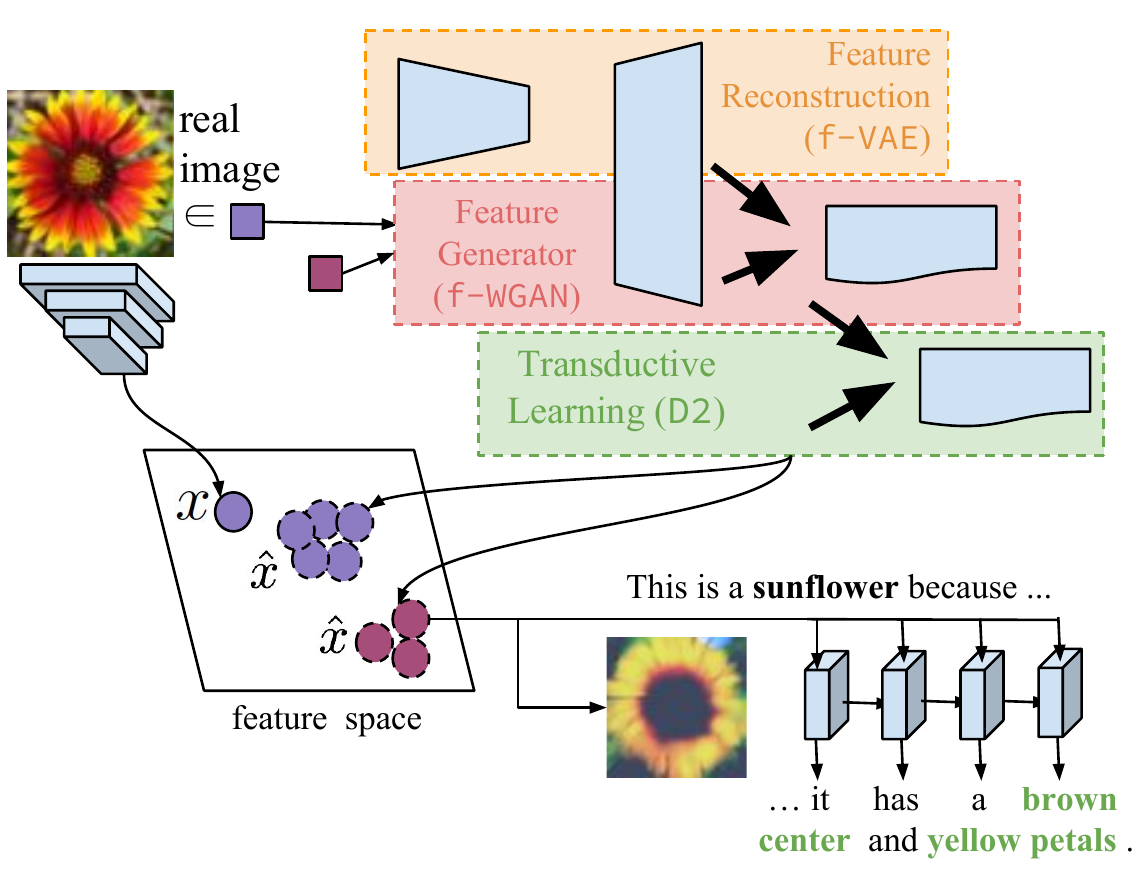}
        \vspace{-5mm}
	\caption{Our any-shot feature generating framework learns discriminative and interpretable CNN features from both labeled data of seen and unlabeled data of novel classes. }
	 \vspace{-3mm}
	\label{fig:teaser}
\end{figure}
 
Our main focus in this work is a new model that generates visual features of any class, utilizing labeled samples when they are available and generalizing to unknown concepts whose labeled samples are not available. Prior work used GANs for this task~\cite{XLSA18,FKRC18} as they directly optimize the divergence between real and generated data, but they suffer from mode collapse issues~\cite{arjovsky2017towards}. On the other hand, feature generation with VAE~\cite{Verma_2018_CVPR} is more stable. However, VAE optimizes the lower bound of log likelihood rather than the likelihood itself~\cite{kingma2013auto}. Our model combines the strengths of VAE and GANs by assembling them to a conditional feature generating model, called \texttt{f-VAEGAN-D2}, that synthesizes CNN image features from class embeddings, i.e. class-level attributes or word2vec~\cite{MSCCD13}. Thanks to its additional discriminator that distinguishes real and generated features, our \texttt{f-VAEGAN-D2} is able to use unlabeled data from previously unseen classes without any condition. 
The features learned by our model, e.g. Figure~\ref{fig:teaser}, are disciminative in that they boost the performance of any-shot learning as well as being visually and textually interpretable.

Our main contributions are as follows. (1) We propose the \texttt{f-VAEGAN-D2} model that consists of a conditional encoder, a shared conditional decoder/generator, a conditional discriminator and a non-conditional discriminator. The first three networks aim to learn the conditional distribution of CNN image features given class embeddings optimizing VAE and WGAN losses on labeled data of seen classes. The last network learns the marginal distribution of CNN image features on the unlabeled features of novel classes. Once trained, our model synthesizes discriminative image features that can be used to augment softmax classifier training. (2) Our empirical analysis on CUB, AWA2, SUN, FLO, and large-scale ImageNet shows that our generated features improve the state-of-the-art in low-shot regimes, i.e. (generalized) zero- and few shot learning in both the inductive and transductive settings. (3) We demonstrate that our generated features are interpretable by inverting them back to the raw pixel space and by generating visual explanations.  

\begin{figure*}[t]
	\centering
        \includegraphics[width=\textwidth, trim=0 0 0 0,clip]{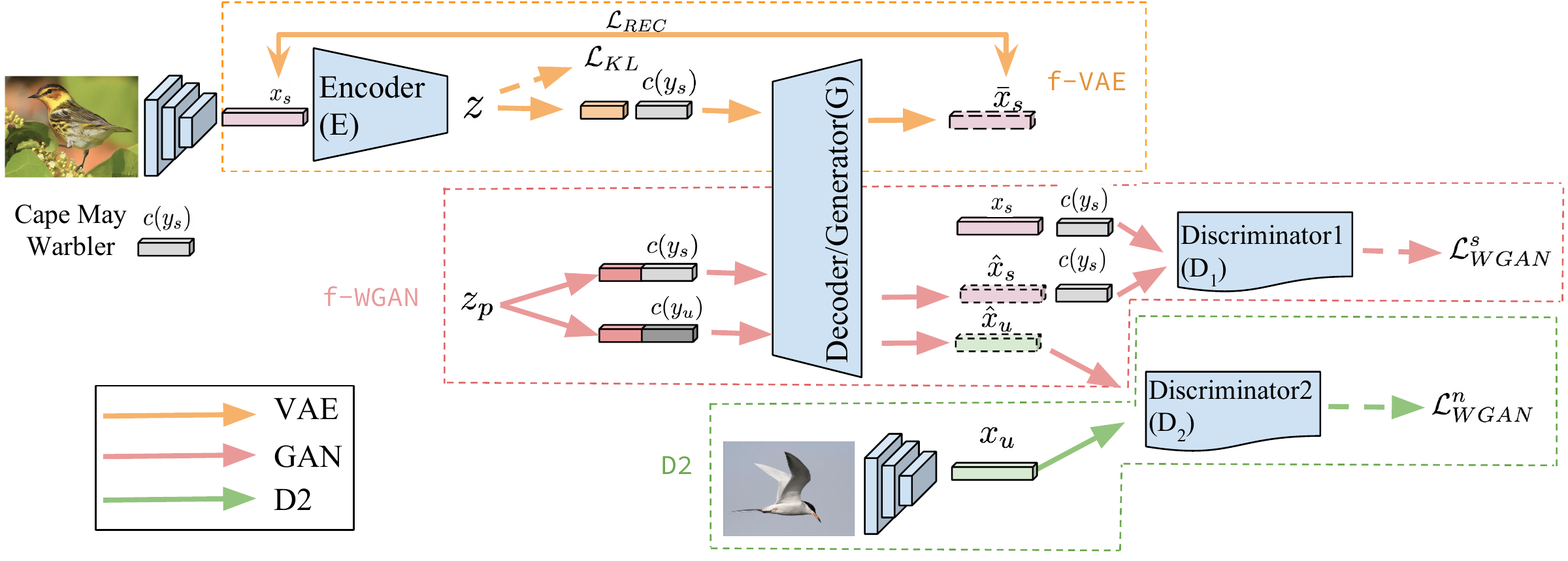}
        \vspace{-5mm}
	\caption{Our any-shot feature generating network (\texttt{f-VAEGAN-D2}) consist of a feature generating VAE (\texttt{f-VAE}), a feature generating WGAN (\texttt{f-WGAN}) with a conditional discriminator~($D_1$) and a transductive feature generator with a non-conditional discriminator~($D_2$) that learns from both labeled data of seen classes and unlabeled data of novel classes.}
	\label{fig:model}
\end{figure*}

\section{Related Work}

In this section, we discuss related works on zero- and few-shot learning as well as generative models.

\myparagraph{Zero-shot Learning.} We are interested in both zero-shot learning~(ZSL) that aims to predict unseen classes and generalized zero-shot learning~(GZSL) that predicts both seen and unseen classes. The required knowledge transfer from seen classes to unseen classes relies on the semantic embedding, e.g. attributes annotated by humans, word embeddings learned on text corpus, hierarchy embeddings obtained from label hierarchy, sentence embeddings from a language model. Unlike the instance-level image features, the semantic embedding is usually class-level, i.e. we use class embedding and semantic embedding interchangeably. 
Early works~\cite{LNH13,jayaraman2014zero} associate seen and unseen classes by learning attribute classifiers. 
Most of recent zero-shot learning works~\cite{APHS13,kodirov2017semantic,RT15,FCSBDRM13,zhang2016learning} learn a compatibility function between the image and semantic embedding spaces.  
\cite{ZV15,NMBSSFCD14,CCGS16} represents image and class embeddings as a mixture of seen class proportions. SYNC~\cite{CCGS16} and \cite{ESE13,lei2015predicting} learn to predict linear classifier weights of unseen classes. \cite{Wang_2018_CVPR} proposes to combine the semantic embedding and knowledge graph with graph convolutional network~\cite{kipf2016semi}. An orthogonal direction is generative model~\cite{verma2017simple, mukherjee2016gaussian}, where class-conditional distribution is learned based on the Gaussian assumption.  

In contrast to those inductive approaches that only use labeled data from seen classes, transductive zero-shot learning methods additionally leverage unlabeled data from unseen classes. PST~\cite{MES13} and DSZSL~\cite{ye2017zero} project image embedding to the semantic embedding space followed by label propagation. TMV~\cite{FHXFG15} combines multiple semantic embeddings and performs hypergraph label propagation. 
\cite{KXFG15, fu2016semi} exploit semantic manifold learning. GFZSL~\cite{verma2017simple} treats unknown labels of unseen class images as latent variables and applies Expectation-Maximization~(EM). As the prediction is biased to seen classes in GZSL, UE~\cite{song2018transductive} maximizes the probability of predicting unlabeled images as unseen classes. Our model operates in both inductive and transductive zero-shot settings. However, unlike most of other transductive approaches that rely on label propagation, we propose to learn a feature generator with labeled data of seen classes and unlabeled data of unseen classes. 

\myparagraph{Few-shot Learning.} The task of few-shot learning is to train a model with only a few training samples. Directly optimizing the standard model with few samples will have high risk of over-fitting. The general idea is to train a model on classes with enough training samples and generalize to classes with few samples without learning new parameters. Siamese neural networks~\cite{koch2015siamese} proposes a CNN architecture that computes similarity between an image pair. Matching network~\cite{vinyals2016matching} and prototypical networks~\cite{snell2017prototypical} predict an image label based on support sets and apply the episode training strategy that mimics the few-shot testing. 
Meta-LSTM~\cite{ravi2016optimization} learns the exact optimization algorithm used to train the few-shot classifier. MAML~\cite{finn2017model} proposes to learn good weight initialization that can be adapted to small dataset efficiently. \cite{HG16,wang2018low} propose a large scale low-shot benchmark on ImageNet and generate features for novel classes.
Imprinting\cite{qi2018low} directly copies the normalized image embedding as classifier weights, while \cite{qiao2018few} predicts classifier weights from image features with a learned neural network. In contrast to those prior works that only rely on visual information, we also leverage class-level semantic information, i.e. attribute or word2vec~\cite{MSCCD13}.

\myparagraph{Generative Models.} Generative modeling aims to learn the probability distribution of data points such that we can randomly sample data from it that can be used as a data augmentation mechanism. Generative Adversarial Networks~(GANs)\cite{GPMXWDOCB14,conditionalgans,RMC16} consist of a generator that synthesizes fake data and a discriminator that distinguishes fake and real data. 
The instable training issues of GANs have been studied by \cite{gulrajani2017improved,arjovsky2017towards,miyato2018spectral}. An interesting application of GANs is CycleGAN~\cite{CycleGAN2017} that translates an image from one domain to another domain. \cite{RAYLSL16} generates natural images from text descriptions, and SRGAN\cite{ledig2017photo} solves single image super-resolution. Variational Autoencoder~(VAE)~\cite{kingma2013auto} employs an encoder that represents the input as a latent variable with Gaussian distribution assumption and a decoder that reconstructs the input from the latent variable. GMMN~\cite{li2015generative} optimizes the maximum mean discrepancy~(MMD)~\cite{gretton2007kernel} between real and generated distribution. Recently, generative models~\cite{BHJ17,Zhu_2018_CVPR, Verma_2018_CVPR,XLSA18} have been applied to solve generalized zero-shot learning by synthesizing CNN features of unseen classes from semantic embeddings. Among those, \cite{BHJ17} uses GMMN~\cite{li2015generative}, 
\cite{Zhu_2018_CVPR, XLSA18} use GANs\cite{GPMXWDOCB14} and \cite{Verma_2018_CVPR} employs VAE~\cite{kingma2013auto}. Our model combines the advantages of both VAE and GAN with an additional discriminator to use unlabeled data of unseen classes which lead to more discriminative features.

\section{\texttt{f-VAEGAN-D2} Model}

Existing models that operate on sparse data regimes are either trained with labeled data from a set of classes which is disjoint from the set of classes at test time, i.e. inductive zero-shot setting~\cite{LNH13,FCSBDRM13}, or the samples can come from all classes but then their labels are not known, i.e. transductive zero-shot setting~\cite{FHXG15,MES13}. Recent works~\cite{XLSA18,Verma_2018_CVPR,FKRC18} address generalized zero-shot learning by generating synthetic CNN features of unseen classes followed by training softmax classifiers, which alleviates the imbalance between seen and unseen classes. However, we argue that those feature generating approaches are not expressive enough to capture complicated feature distributions in real world. In addition, since they have no access to any real unseen class features, there is no guarantee on the quality of generated unseen class features. As shown in Figure~\ref{fig:model}, we proposes to enhance the feature generator by combining VAE and GANs with shared decoder and generator, and adding another discriminator~($D_2$) to distinguish real or generated features without applying any condition. Intuitively, in transductive zero-shot setting, by feeding real unlabeled features of unseen classes, $D_2$ will be able to learn the manifold of unseen class such that more realistic features can be generated.
Hence, the key to our approach is the ability to generate semantically rich CNN feature distributions, which is generalizes to any-shot learning scenarios ranging from (generalized) zero-shot to (generalized) few-shot to (generalized) many-shot learning.

\myparagraph{Setup.}
We are given a set of images $X=\{x_1,\ldots,x_l\} \cup \{x_{l+1},\ldots,x_t \}$ encoded in the image feature space $\mathcal{X}$, a seen class label set $Y^s$, a novel label set $Y^n$, a.k.a unseen class label set $Y^u$ in the zero-shot learning literature. The set of class embeddings  $C=\{c(y)| \forall y \in Y^s \cup Y^n\}$ are encoded in the semantic embedding space $\mathcal{C}$ that defines high level semantic relationships between classes. The first $l$ points $x_s(s\leq l)$ are labeled as one of the seen classes $y_s\in Y^s$ and the remaining points $x_n(l+1\leq n \leq t)$ are unlabeled, i.e. may come from seen or novel classes. 

In the inductive setting, the training set contains only labeled samples of seen class images, i.e. $\{x_1,\ldots,x_l\}$. On the other hand, in the transductive setting, the training set contains both labeled and unlabeled samples, i.e. $\{x_1,\ldots,x_l,x_{l+1},\ldots,x_t \}$. For both inductive and transductive settings the inference is the same. In zero-shot learning, the task is to predict the label of those unlabeled points that belong to novel classes, i.e. $f_{zsl}: \mathcal{X} \rightarrow \mathcal{Y}^n$, while in the generalized zero-shot learning, the goal is to classify those unlabeled points that can be either from seen or novel classes, i.e. $f_{gzsl}: \mathcal{X} \rightarrow \mathcal{Y}^s \cup \mathcal{Y}^n$. Few-shot and generalized few-shot learning are defined similarly. 

Our framework can be thought of as a data augmentation scheme where arbitrarily many synthetic features of sparsely populated classes aid in improving the disciminative power of classifiers. In the following, we only detail our feature generating network structure as the classifier is unconstrained (we use linear softmax classifiers).

\subsection{Baseline Feature Generating Models}
In feature generating networks (\texttt{f-WGAN})~\cite{XLSA18} the generator $G(z,c)$ generates a CNN feature $\hat{x}$ in the input feature space $\mathcal{X}$ from random noise $z_p$ and a condition $c$, and the discriminator $D(x,c)$ takes as input a pair of input features $x$ and a condition $c$ and outputs a real value, optimizing:
\begin{align}
\label{eq:wgan}
\mathcal{L}^s_{WGAN} =& \mathbb{E}[D(x,c)] - \mathbb{E}[D(\tilde{x},c)] \\ 
                   & - \lambda \mathbb{E}[\left(||\nabla_{\hat{x}} D(\hat{x},c)||_2 - 1\right)^2], \nonumber
\end{align}
where $\tilde{x}=G(z, c)$ is the generated feature and $\hat{x} = \alpha x +(1-\alpha x)$ with $\alpha \sim U(0,1)$ and $\lambda$ is the penalty coefficient. 

The feature generating VAE~\cite{kingma2013auto} (\texttt{f-VAE}) consists of an encoder $E(x,c)$, which encodes an input feature $x$ and a condition $c$ to a latent variable $z$, and a decoder $Dec(z,c)$, which reconstructs the input $x$ from the latent $z$ and condition $c$ optimizing:  
\begin{align}
\label{eq:kl}
\mathcal{L}^s_{VAE} & = KL(q(z|x, c)||p(z|c))  \\ 
& -\mathbb{E}_{q(z|x,c)}[\log p(x|z,c)], \nonumber
\end{align}
where the conditional distribution $q(z|x, c)$ is modeled as $E(x,c)$, $p(z|c))$ is assumed to be $\mathcal{N}(0,1)$, KL is the Kullback-Leibler divergence, and $p(x|z,c)$ is equal to $Dec(z,c)$. 

\subsection{Our \textbf{\texttt{f-VAEGAN-D2}} Model}
It has been shown that ensembling a VAE and a GAN leads to better image generation results~\cite{larsen2015autoencoding}. We hypothesize that VAE and GAN learn complementary information for feature generation as well. This is likely when the target data follows a complicated multi-modal distribution where two losses are able to capture different modes of the data.

To combine \texttt{f-VAE} and \texttt{f-WGAN}, we introduce an encoder $E(x,c): \mathcal{X} \times \mathcal{C}\rightarrow \mathcal{Z}$, which encodes a pair of feature and class embedding to a latent representation, and a discriminator $D_1: \mathcal{X} \times \mathcal{C} \rightarrow \mathbb{R}$ maps this embedding pair to a compatibility score, optimizing: 
\begin{align}
\label{eq:tran}
\mathcal{L}^s_{VAEGAN} = \mathcal{L}^s_{VAE} + \gamma \mathcal{L}^s_{WGAN}
\end{align}
where the generator $G(z, c)$ of the GAN and decoder $Dec(z, c)$ of the VAE share the same parameters.  
The superscript $s$ indicates that the loss is applied to feature and class embedding pair of seen classes. $\gamma$ is a hyperparameter to  control the weighting of VAE and GAN losses.

Furthermore, when unlabeled data of novel classes becomes available, we propose to add a non-conditional discriminator $D_2$ (\texttt{D2} in \texttt{f-VAEGAN-D2}) which distinguishes between real and generated features of novel classes. This way $D_2$ learns the feature manifold of novel classes. 
Formally, our additional non-conditional discriminator $D_2: \mathcal{X} \rightarrow R$ distinguishes real and synthetic unlabeled samples using a WGAN loss:
\begin{align}
\label{eq:tran}
\mathcal{L}^n_{WGAN} =& \mathbb{E}[D_2(x_n)] - \mathbb{E}[D_2(\tilde{x}_n)] - \\ 
& \lambda \mathbb{E}[\left(||\nabla_{\hat{x}_n}D_2(\hat{x}_n)||_2 - 1\right)^2], \nonumber
\end{align}
where $\tilde{x}_n = G(z, y_n)$ with $y_n\in Y^n$, $\hat{x}_n = \alpha x_n +(1-\alpha x_n)$ with $\alpha \sim U(0,1)$. Since $\mathcal{L}^s_{WGAN}$ is trained to learn CNN features using labeled data conditioned on class embeddings of seen classes and class embeddings encode shared properties across classes, we expect these CNN features to be transferable across seen and novel classes. However, this heavily relies on the quality of semantic embeddings and suffers from domain shift problems. Intuitively, $\mathcal{L}^n_{WGAN}$  captures the marginal distribution of CNN features and provides useful signals of novel classes to generate transferable CNN features. Hence, our unified \texttt{f-VAEGAN-D2} model optimizes the following objective function: 
\begin{align}
\min_{G,E} \max_{D_1, D_2} \mathcal{L}^s_{VAEGAN} + \mathcal{L}^n_{WGAN}
\end{align}

\myparagraph{Implementation Details.}
Our generator ($G$) and discriminators ($D_1$ and $D_2$) are implemented as multilayer perceptron~(MLP). The random Gaussian noise $z\sim N(0, 1)$ and class embedding $c(y)$ are concatenated and fed into the generator, which is composed of 2 fully connected layers with 4096 hidden units. We find dimension of noise $d_z=d_c$, i.e. dimension of class embeddings, works well. Similarly, the discriminators take input as the concatenation of image feature and class embedding and have 2 fully connected layers with 4096 hidden units. We use LeakyReLU as the nonlinear activation function except for the output layer of $G$, for which Sigmoid is used because we apply binary cross entropy loss as $\mathcal{L}_{REC}$ and input features are rescaled to be in $[0,1]$. We find $\beta=1$ and $\gamma=1000$ works well across all the datasets. Gradient penalty coefficient is set to $\lambda=10$ and generator is updated every 5 discriminator iterations as suggested in WGAN paper~\cite{arjovsky2017wasserstein}. As for the optimization, we use Adam optimizer with constant learning rate $0.001$ and early stopping on the validation set.

\section{Experiments}
In this section, we validate our approach in both zero-shot and few-shot learning. The details of the settings are provided in their respective sections.

{
\setlength{\tabcolsep}{6pt}
\renewcommand{\arraystretch}{1.2} 
\begin{table}[t]
 \centering
   \begin{tabular}{l l c c}
    & \textbf{Model} & \textbf{ZSL} & \textbf{GZSL} \\ \hline
    \multirow{3}{*}{INDUCTIVE} & \texttt{GAN} & $59.1$ & $52.3$ \\
    & \texttt{VAE} & $58.4$ & $52.5$ \\
    & \texttt{VAE-GAN} & $61.0$ & $53.7$ \\
    \hline
    \multirow{3}{*}{TRANSDUCTIVE} & \texttt{GAN} & $67.3$ & $61.6$ \\
    & \texttt{VAE} & $68.9$ & $59.6$ \\
    & \texttt{VAE-GAN} & $\mathbf{71.1}$ & $\mathbf{63.2}$
     \end{tabular}
\caption{Ablating different generative models on CUB (using attribute class embedding and image features with no fine-tuning). ZSL: top-1 accuracy on unseen classes, GZSL: harmonic mean of seen and unseen class accuracies. }
\label{tab:ablation}
\end{table}
}

{
\setlength{\tabcolsep}{4pt}
\renewcommand{\arraystretch}{1.2} 
\begin{table*}[t]
 \centering
 \resizebox{\linewidth}{!}{%
   \begin{tabular}{l l c c c c |c c c |c c c |c c c | c c c  }
     & & \multicolumn{4}{c|}{\textbf{Zero-Shot Learning}} & \multicolumn{12}{c}{\textbf{Generalized Zero-Shot Learning}} \\
     & & \textbf{CUB} & \textbf{FLO} & \textbf{SUN} &  \textbf{AWA} & \multicolumn{3}{c}{\textbf{CUB}} & \multicolumn{3}{c}{\textbf{FLO}} & \multicolumn{3}{c}{\textbf{SUN}} & \multicolumn{3}{c}{\textbf{AWA}}    \\
     & Method & \textbf{T1} & \textbf{T1} & \textbf{T1} & \textbf{T1}&  \textbf{u} & \textbf{s} & \textbf{H} & \textbf{u} & \textbf{s} & \textbf{H} & \textbf{u} & \textbf{s} & \textbf{H} & \textbf{u} & \textbf{s} & \textbf{H}  \\
     \hline
     \multirow{6}{*}{IND} & \texttt{ALE}~\cite{APHS15} & $54.9$ & $48.5$ & $58.1$ & $59.9$
     & $23.7$ & $62.8$ & $34.4$ & $13.3$ & $61.6$ & $21.9$ & $21.8$ & $33.1$ & $26.3$ & $16.8$ & $76.1$ & $27.5$ \\ 
     & \texttt{CLSWGAN}~\cite{XLSA18} &  $57.3$ & $67.2$ & $60.8$ & $68.2$ & $43.7$ & $57.7$ & $49.7$ & $59.0$ & $73.8$ & $65.6$ &  $42.6$ & $36.6$ & $39.4$ & $57.9$ & $61.4$ & $59.6$ \\  
     & \texttt{SE-GZSL}~\cite{Verma_2018_CVPR} &  $59.6$ & - & $63.4$ & $69.2$ & $41.5$ & $53.3$ & $46.7$ & - & - & - &  $40.9$ & $30.5$ & $34.9$ & $58.3$ & $68.1$ & $62.8$ \\  
     & \texttt{Cycle-CLSWGAN}~\cite{FKRC18} & $58.6$ & $70.3$ & $59.9$ & $66.8$ & $47.9$ & $59.3$ & $53.0$ & $61.6$ & $69.2$ &  $65.2$ & $47.2$ & $33.8$ & $39.4$ & $\mathbf{59.6}$ & $63.4$ & $59.8$ \\
     & \texttt{Ours} &  $61.0$ & $67.7$ & $64.7$ & $\mathbf{71.1}$ & $48.4$ & $60.1$ & $53.6$ & $56.8$ & $74.9$ & $64.6$ &  $45.1$ & $\mathbf{38.0}$ & $41.3$ & $57.6$ & $70.6$ & $63.5$ \\    
     & \texttt{Ours-finetuned} & $\mathbf{72.9}$ & $\mathbf{70.4}$ & $\mathbf{65.6}$ & $70.3$ & $\mathbf{63.2}$ & $\mathbf{75.6}$  & $\mathbf{68.9}$ & $\mathbf{63.3}$ & $\mathbf{92.4}$ & $\mathbf{75.1}$ & $\mathbf{50.1}$ & $37.8$ & $\mathbf{43.1}$ & $57.1$ & $\mathbf{76.1}$ & $\mathbf{65.2}$ \\
     \hline
     \multirow{6}{*}{TRAN} & \texttt{ALE-tran}~\cite{xian2018zero} & $54.5$ & $48.3$ & $55.7$ & $70.7$
     & $23.5$ & $45.1$ & $30.9$ & $13.6$ & $61.4$ & $22.2$ & $19.9$ & $22.6$ & $21.2$ & $12.6$ & $73.0$ & $21.5$ \\
     & \texttt{GFZSL}~\cite{verma2017simple} &  $50.0$ & $85.4$ & $64.0$ & $78.6$ & $24.9$ & $45.8$ & $32.2$ & $21.8$ & $75.0$ & $33.8$ &  $0.0$ & $41.6$ & $0.0$ & $31.7$ & $67.2$ & $43.1$ \\  
      & \texttt{DSRL}~\cite{ye2017zero} &  $48.7$ & $57.7$ & $56.8$ & $72.8$ & $17.3$ & $39.0$ & $24.0$ & $26.9$ & $64.3$ & $37.9$ &  $17.7$ & $25.0$ & $20.7$ & $20.8$ & $74.7$ & $32.6$ \\ 
      & \texttt{UE-finetune}~\cite{song2018transductive} &  $72.1$ & - & $58.3$ & $79.7$ & $74.9$  & $71.5$  & $73.2$  & - & - & -  & $33.6$  & $\mathbf{54.8}$ & $41.7$ & $\mathbf{93.1}$ & $66.2$  & $77.4$  \\    
    & \texttt{Ours} & $71.1$ & $89.1$ & $70.1$ & $\mathbf{89.8}$ & $61.4$ & $65.1$ & $63.2$ & $78.7$ & $87.2$ & $82.7$ & $\mathbf{60.6}$ & $41.9$ & $\mathbf{49.6}$ & $84.8$ & $88.6$  & $86.7$ \\    
    & \texttt{Ours-finetuned} & $\mathbf{82.6}$ & $\mathbf{95.4}$ &  $\mathbf{72.6}$ & $89.3$ & $\mathbf{73.8}$ & $\mathbf{81.4}$  & $\mathbf{77.3}$ &  $\mathbf{91.0}$ & $\mathbf{97.4}$ & $\mathbf{94.1}$ & $54.2$ & $41.8$ & $47.2$ & $86.3$ & $\mathbf{88.7}$ & $\mathbf{87.5}$
\end{tabular}
   } \vspace{-2mm}
\caption{Comparing with the-state-of-the-art. Top: inductive methods (IND), Bottom: transductive methods (TRAN). Fine tuning is performed only on seen class images as this does not violate the zero-shot condition. We measure top-1 accuracy (\textbf{T1}) in ZSL setting, Top-1 accuracy on seen (\textbf{s}) and unseen (\textbf{s}) classes as well as their harmonic mean (\textbf{H}) in GZSL setting.
}
\label{tab:zsl_main}
\end{table*}
}

\subsection{(Generalized) Zero-shot Learning}
We validate our model on five widely-used datasets for zero-shot learning, i.e. Caltech-UCSD-Birds~(CUB)~\cite{CaltechUCSDBirdsDataset}, Oxford Flowers~(FLO)~\cite{OxfordFlowersDataset}, SUN Attribute~(SUN)~\cite{PH12} and Animals with Attributes2~(AWA2)~\cite{xian2018zero}. Among those, CUB, FLO and SUN are medium scale, fine-grained datasets. AWA2, on the other hand, is a coarse-grained dataset. Finally we evaluate our model also on ImageNet~\cite{imagenet} with more than 14 million images and 21K classes as a large-scale and fine-grained dataset. 

We follow the exact ZSL and GZSL splits as well as the evaluation protocol of~\cite{xian2018zero} and for fair comparison we use the same image and class embeddings for all models. Briefly, image (with no image cropping or flipping) features are extracted from the 2048-dim top pooling units of 101-layer ResNet pretrained  on ImageNet 1K. For comparative studies, we also fine-tune ResNet-101 on the seen class images of each dataset. As for class embeddings, unless otherwise specified, we use class-level attributes for CUB~(312-dim), AWA2~(85-dim) and SUN(102-dim). For CUB and FLO, we also extract 1024-dim sentence embeddings of character-based CNN-RNN model~\cite{RALS16} from fine-grained visual descriptions (10 sentences per image). 

\myparagraph{Ablation study.} We ablate our model with respect to the generative model, i.e. using \texttt{GAN}, \texttt{VAE} or \texttt{VAE-GAN} in both inductive and transductive settings. Our conclusions from Table~\ref{tab:ablation}, are as follows. In the inductive setting \texttt{VAE-GAN} has an edge over both \texttt{VAE} and \texttt{GAN}, i.e. $59.1\%$  and $58.4\%$  vs $61.0\%$ in ZSL setting. Adding unlabeled samples to the training set, i.e. transductive learning setting, is beneficial for all the generative models. As in the inductive setting \texttt{VAE} and \texttt{GAN} achieve similar results, i.e $67.3\%$ and $68.9\%$ for ZSL. Our \texttt{VAE-GAN} model leads to the state-of-the-art results, i.e. $71.1\%$ in ZSL and $63.2\%$ in GZSL confirming that VAE and GAN learn complementary representations. As \texttt{VAE-GAN} gives the highest accuracy in all settings, it is employed in all remaining results of the paper.

\begin{figure}[t]
	\centering
		\includegraphics[width=.48\columnwidth, trim=10 10 50 0,clip]{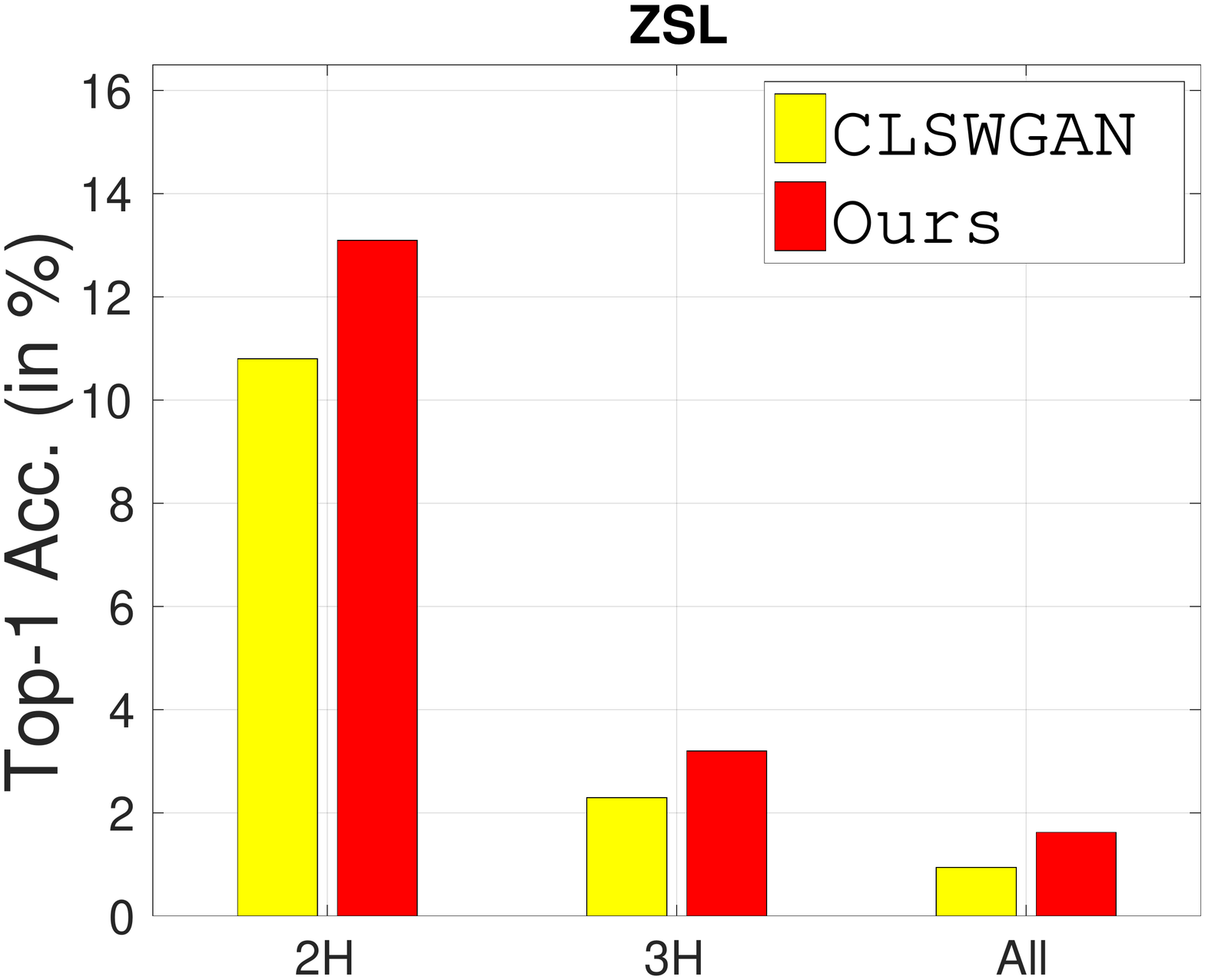}
        \includegraphics[width=.48\columnwidth, trim=10 10 50 0,clip]{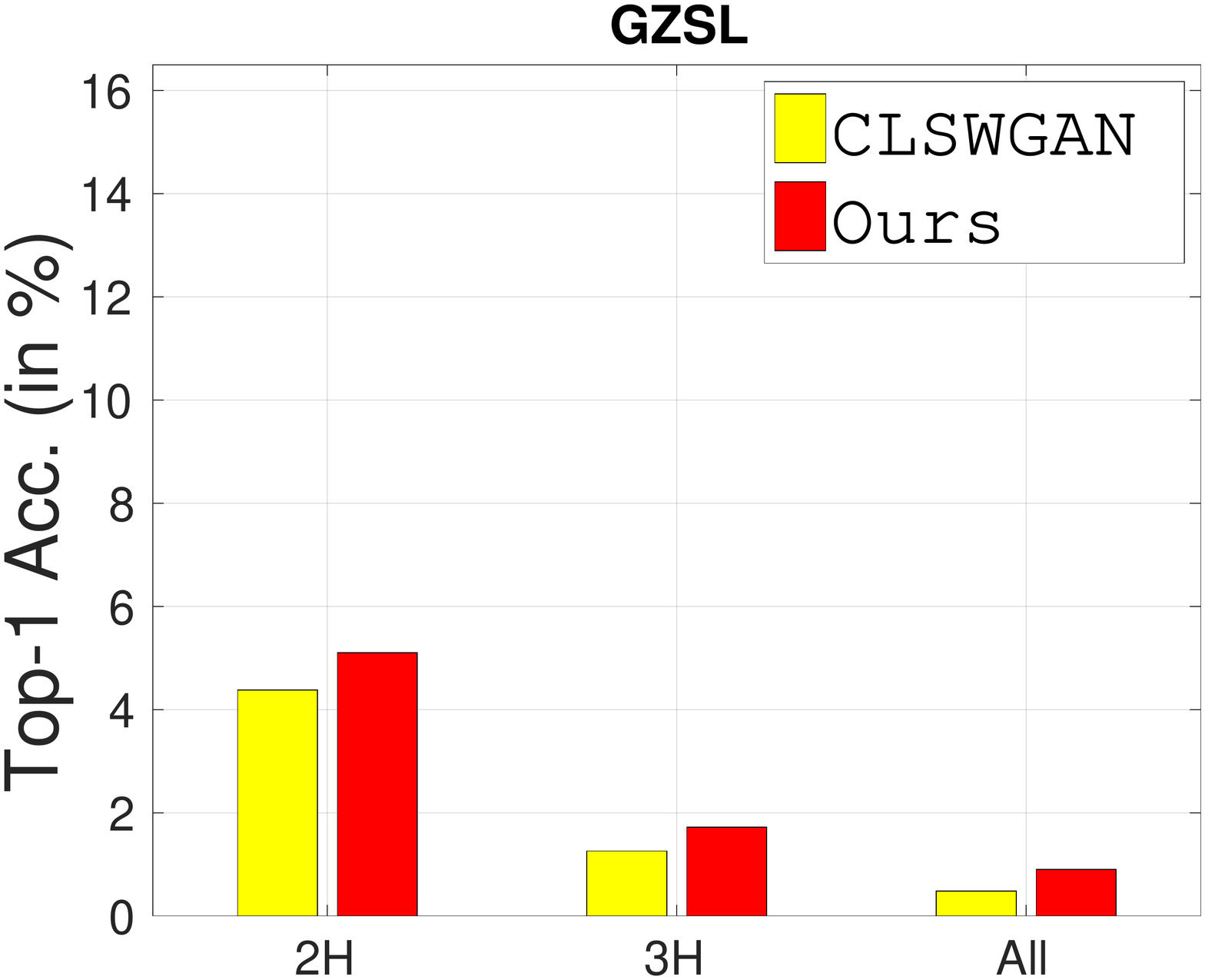}
	\caption{Top-1 ZSL results on ImageNet. We follow the splits in~\cite{xian2018zero} and compare our results with the state-of-the-art feature generating model \texttt{CLSWGAN}~\cite{XLSA18}.}
	\label{fig:imagenet}
\end{figure}

\begin{figure*}[t]
	\centering
    \begin{subfigure}[b]{0.49\textwidth}
        \includegraphics[width=0.49\textwidth, trim=5 0 25 0,clip]{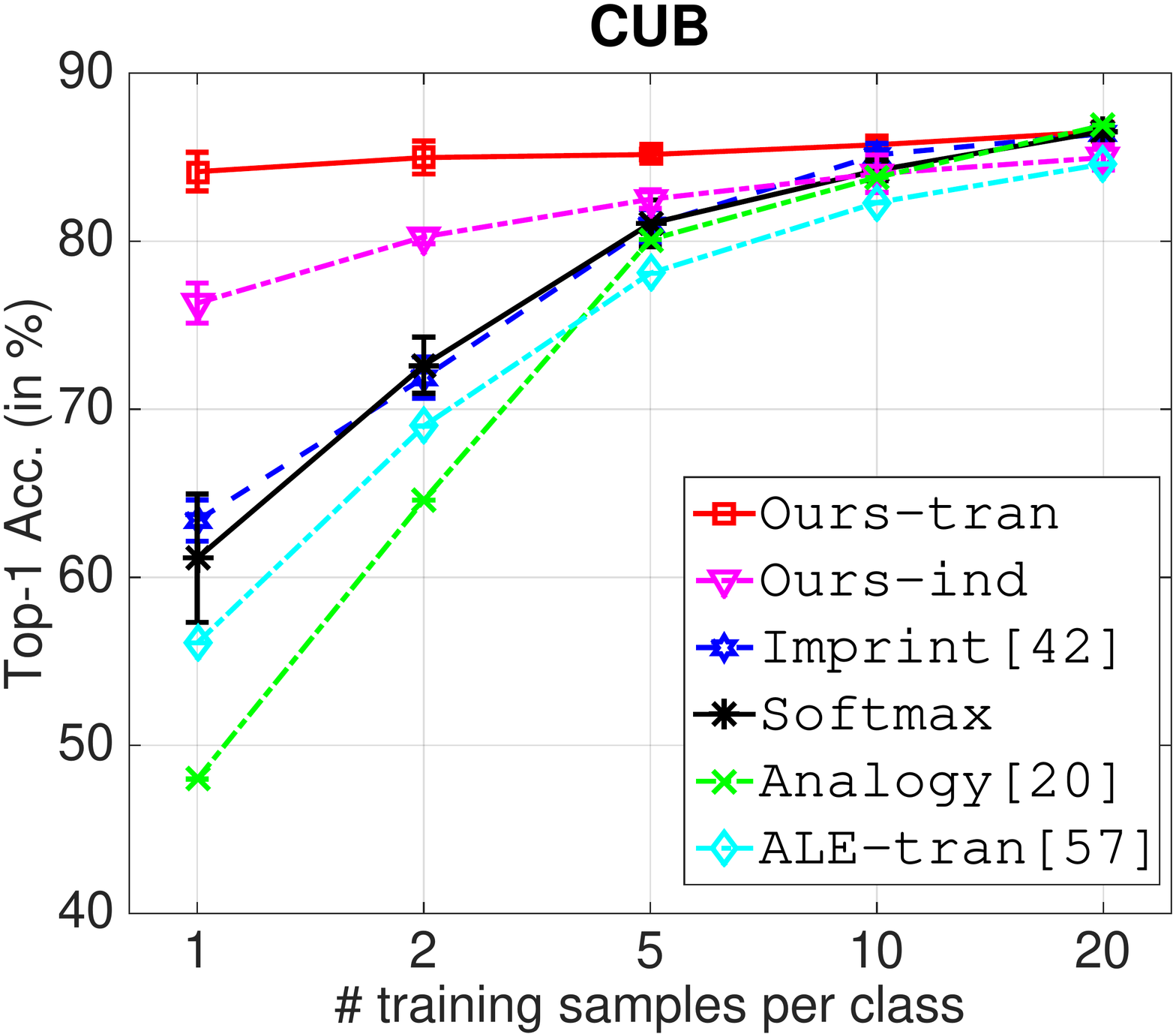} 
        \includegraphics[width=0.49\textwidth, trim=5 0 25 0,clip]{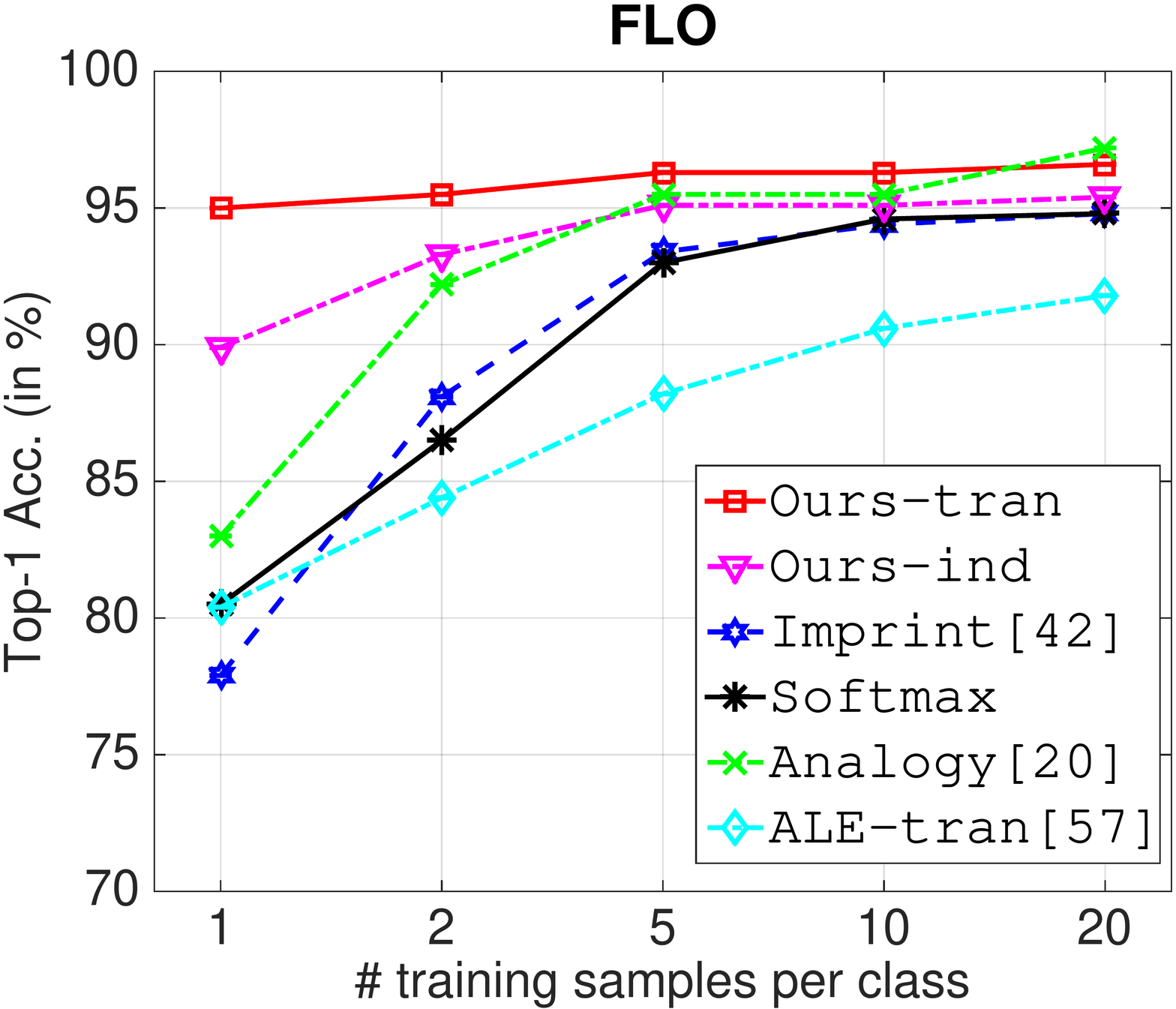} 
        \caption{Few-Shot Learning (FSL)}
        \end{subfigure}
     \begin{subfigure}[b]{0.49\textwidth}
        \includegraphics[width=.49\textwidth, trim=5 0 25 0,clip]{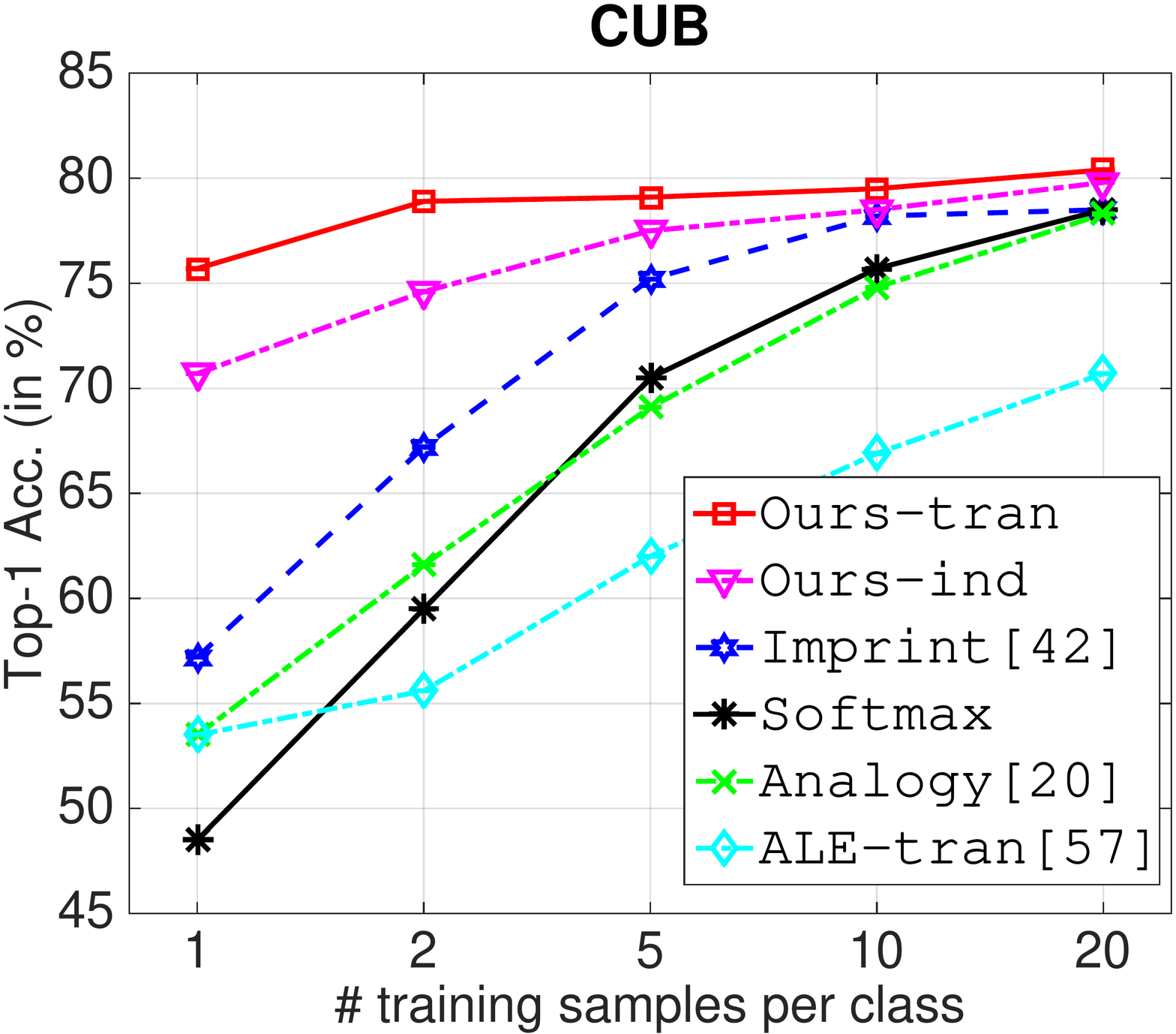} 
        \includegraphics[width=.49\textwidth, trim=5 0 25 0,clip]{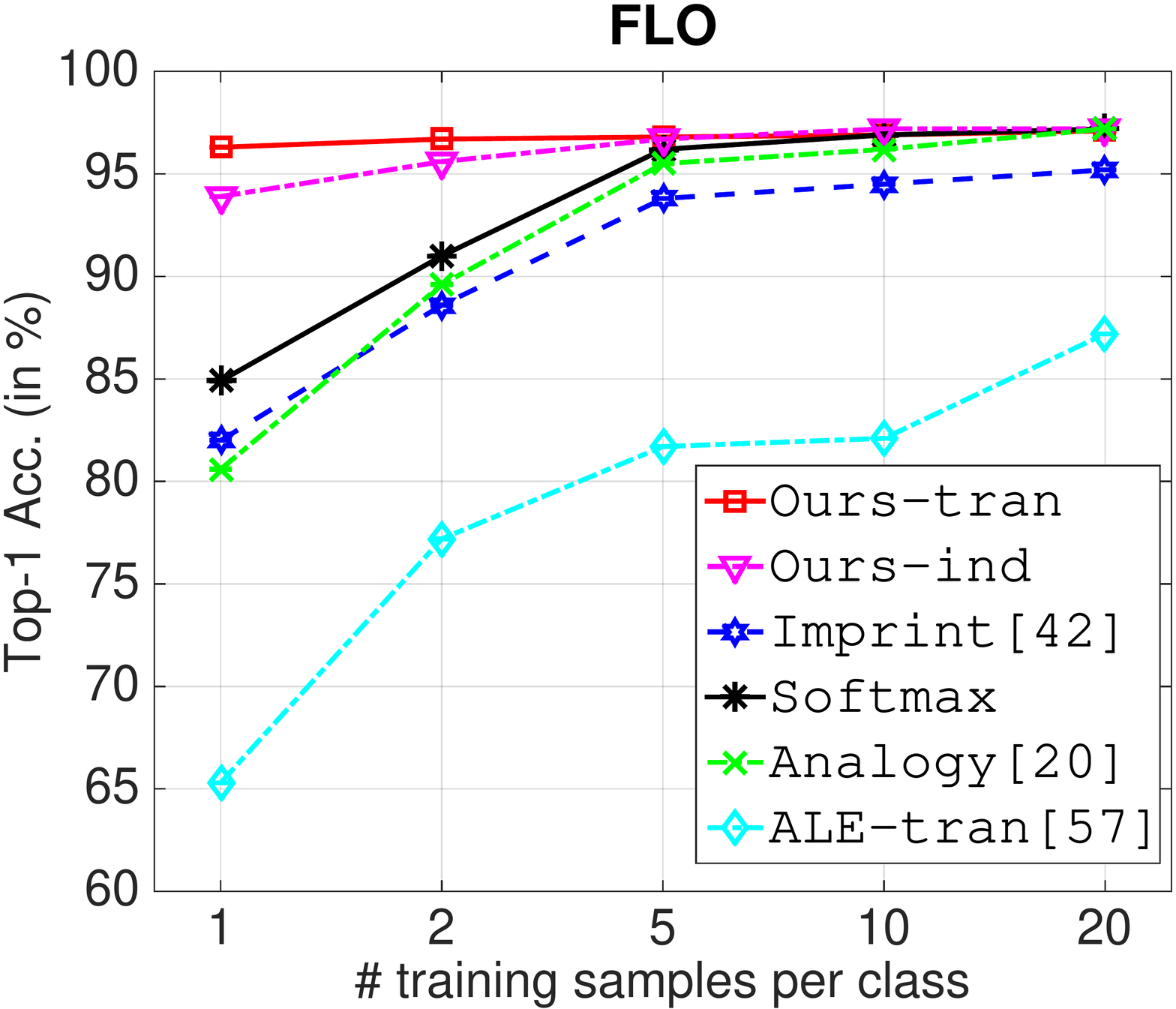} 
        \caption{Generalized Few-Shot Learning (GFSL)}
        \end{subfigure}
    	\caption{FSL and GFSL results on CUB and FLO with increasing number of training samples per novel class. Left: FSL plots show the top-1 accuracy on novel classes. Right: GZSL plots show the top-1 accuracy on all classes.} 
	\label{fig:fewshot}
\end{figure*}

\myparagraph{Comparing with the state-of-the-art.} In Table~\ref{tab:zsl_main} we compare our model with the best performing recent methods on four zero-shot learning datasets on ZSL and GZSL settings. 

In the inductive ZSL setting, our model both with and without fine-tuning outperforms the state-of-the art for all datasets. Our model with fine-tuned features establishes the new state-of-the-art, i.e. $72.9\%$ on CUB, $70.4\%$ on FLO, $65.6\%$ on SUN and $70.3\%$ on AWA. For the transductive ZSL setting, our model without fine-tuning on CUB is surpassed by UE-finetune of~\cite{song2018transductive}, i.e. $71.1\%$ vs $72.1\%$. However, when we also fine-tune our features, we establish the new state-of-the-art on the transductive ZSL setting as well, i.e. $82.6\%$ on CUB, $95.4\%$ on FLO, $72.6\%$ on SUN and $89.3\%$ on AWA.

In the GZSL setting, we observe that feature generating methods, i.e. our model, \texttt{CLSWGAN}~\cite{XLSA18}, \texttt{SE-GZSL}~\cite{Verma_2018_CVPR}, \texttt{Cycle-CLSWGAN}~\cite{FKRC18} achieve better results than others. This is due to the fact that data augmentation through feature generation leads to a more balanced data distribution such that the learned classifier is not biased to seen classes. Note that although \texttt{UE}~\cite{song2018transductive} is not a feature generating method, it leads to strong results as this model uses additional information, i.e. it assumes that unlabeled test samples always come from unseen classes. Nevertheless, our model with fine-tuning leads to $77.3\%$ harmonic mean~(H) on CUB, $94.1\%$ H on FLO, $47.2\%$ H on SUN and $87.5\%$ H on AWA achieving significantly higher results than all the prior works.

\myparagraph{Large-scale experiments.} Although most of the prior work presented in Table~\ref{tab:zsl_main} has not been evaluated in ImageNet, this dataset serves a challenging and interesting test bed for (G)ZSL research. Hence, we compare our model with \texttt{CLSWGAN}~\cite{XLSA18} on ImageNet using the same evaluation protocol. As shown in Figure~\ref{fig:imagenet} our model significantly improves over the state-of-the-art in both ZSL and GZSL settings in 2H, 3H and All splits determined by considering the classes 2 hops or 3 hops away from 1000 classes of Imagenet as well as all the remaining classes. These experiments are important for two reasons. First, they show that our feature generation model is scalable to the largest scale setting available. Second, our model is applicable to the situations even when human annotated attributes are not available, i.e. for ImageNet classes attributes are not available hence we use per-class word2vec representations.

\subsection{(Generalized) Few-shot Learning}
In few-shot or low-shot learning scenarios, classes are divided into base classes that have a large number of labeled training samples and novel classes that contain only few labeled samples per category. In the plain FSL setting, the goal is to achieve good performance on novel classes whereas in GFSL setting good performance must generalize to all classes. 

Among the classic ZSL datasets, CUB has been used for few-shot learning in~\cite{qi2018low} by taking the first 100 classes as base classes and the rest as novel classes. However, as ImageNet 1K contains some of those novel classes and feature extractors are pretrained on it, we use the class splits from the standard ZSL setting, i.e. 150 base and 50 novel. For FLO we also follow the same class splits as in ZSL. As for features, we use the same fine-tuned ResNet-101 features and attribute class embeddings used in zero-shot learning experiments. For fairness, we repeat all the experiments for \cite{qi2018low} and \cite{HG16} with the same image features.

\myparagraph{Comparing with the state-of-the-art.} As shown in Figure~\ref{fig:fewshot} both for FSL and GFSL settings and for both datasets both our inductive and transductive models have a significant edge over all the competing methods when the number of samples from novel classes is small, e.g. 1,2 and 5. This shows that our model generates highly discriminative features even with only few real samples are present. In fact, only with one real sample per class, our model achieves almost the full accuracy obtained with 20 samples per class. Going towards the full supervised learning, e.g. with 10 or 20 samples per class, all methods perform similarly. This is expected since in the setting where a large number of labeled samples per class is available, then a simple softmax classifier that uses real ResNet-101 features achieves the state-of-the-art.

In the inductive FSL setting, our model that uses one labeled sample per class reaches the accuracy as softmax that uses five samples per class. In the transductive FSL setting, our model that uses one labeled sample per class reaches the accuracy of softmax obtained with 10 samples per class. Furthermore, the inductive GFSL setting, our model with two samples per class achieves the same accuracy as softmax trained with ten samples per class on CUB. In the transductive GFSL setting, for FLO, for our model only one labeled sample is enough to reach the accuracy obtained with 20 labeled samples with softmax. Note that the same behavior is observed on SUN and AWA as well. Due to space restrictions we present them in the supplementary material.

\paragraph{Large-scale experiments.} 
Regarding few-shot learning results on ImageNet, we follow the procedure in~\cite{HG16} where 1K ImageNet categories are randomly divided into 389 base and 611 novel classes. To facilitate cross validation, base classes are further split into $C^1_{base}$ (193 classes) and $C^2_{base}$ (196 classes), and novel classes into $C^1_{novel}$ (300 classes) and $C^2_{novel}$ (311 classes). The cross validation of hyperparameters is performed on $C^1_{base}$  and $C^1_{novel}$ and the final results are reported on $C^2_{base}$  and $C^2_{novel}$. Here, we extract image features from the ResNet-50 pretrained on $C^1_{base}\cup C^2_{base}$, which is provided by the benchmark~\cite{HG16}. Since there is no attribute annotation on ImageNet, we use 300-dim word2vec~\cite{MSCCD13} embeddings as the class embedding. Following \cite{wang2018low}, we measure the averaged top-5 accuracy on test examples of novel classes with the model restricted to only output novel class labels, and the averaged top-5 accuracy on test examples of all classes with the model that predicts both base and novel classes. 

\begin{figure}[t]
	\centering
		\includegraphics[width=.48\columnwidth, trim=10 0 50 0,clip]{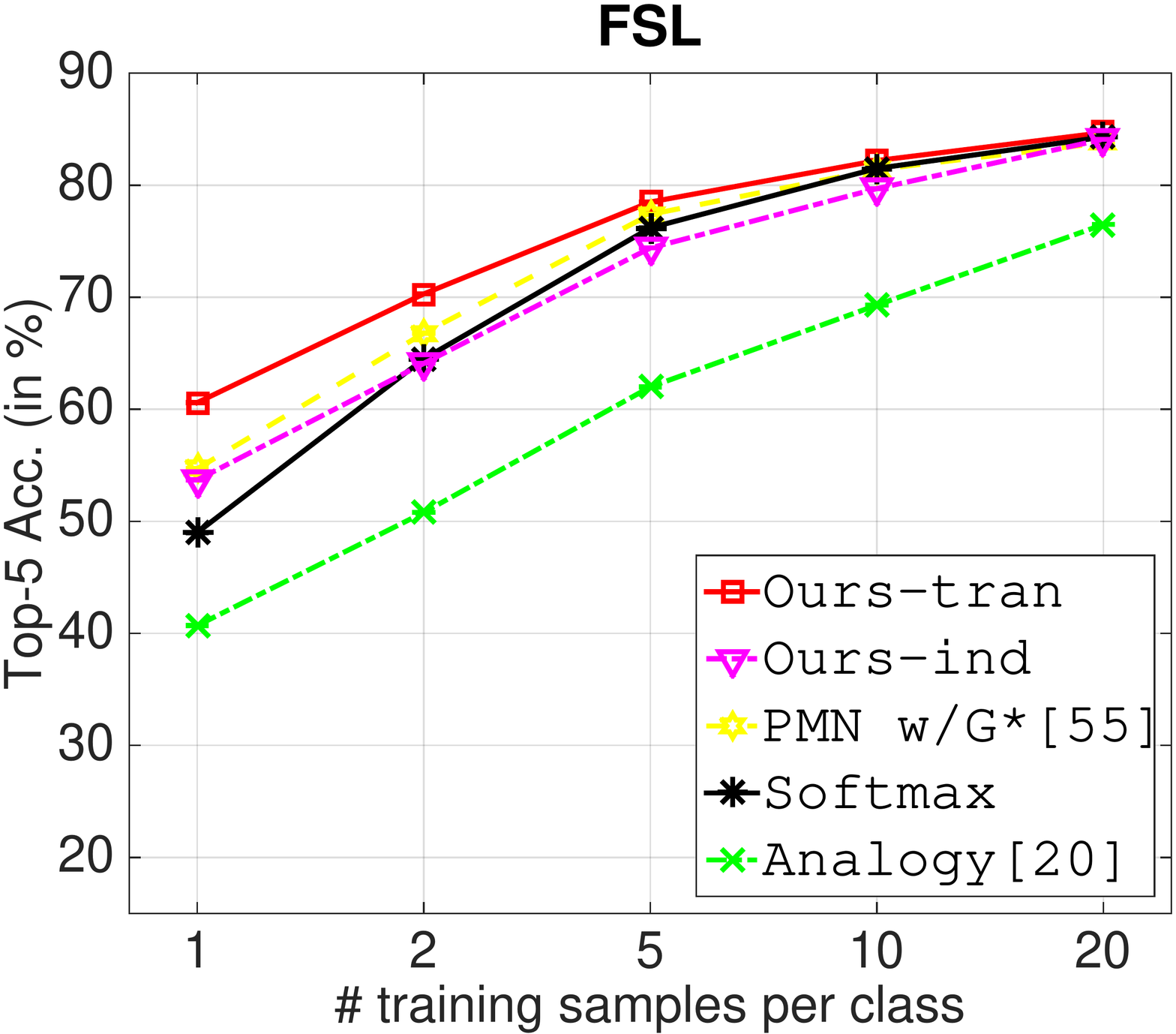}
        \includegraphics[width=.48\columnwidth, trim=10 0 50 0,clip]{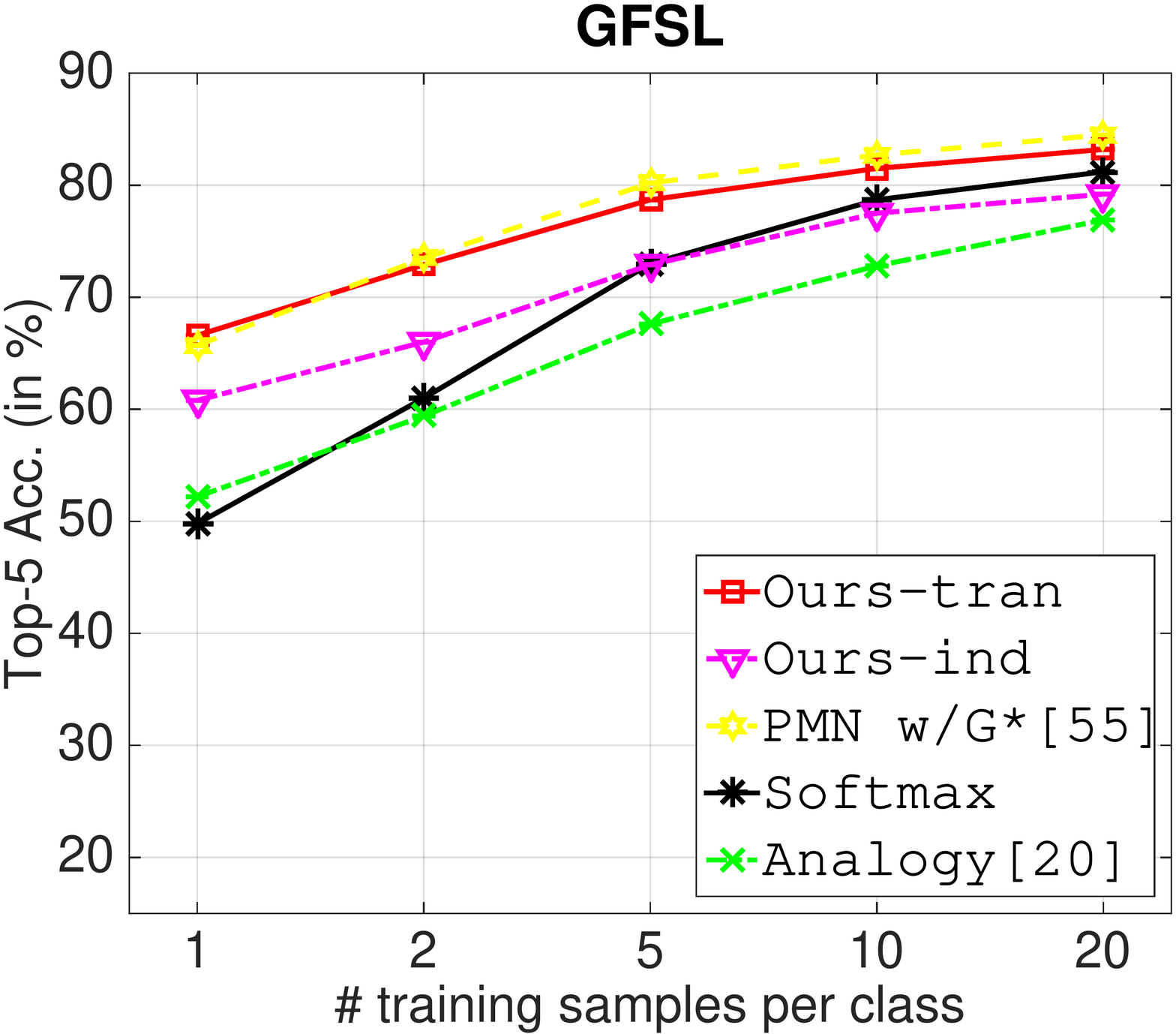}
	\caption{Few Shot Learning results on ImageNet with increasing number of training samples per novel class (Top-5 Accuracy). Left: FSL setting, Right: GFSL setting.}
	\label{fig:imagenet_fsl}
\end{figure}

Our baselines are PMN w/G*~\cite{wang2018low} combining meta-learning and feature generation, analogy generator~\cite{HG16} learning an analogy-based feature generator and softmax classifier learned with uniform class sampling. For, few-shot learning results in Figure~\ref{fig:imagenet_fsl}(left), we observe that our model in the transductive setting, i.e. \texttt{Ours-tran} improves the state-of-the-art PMN w/G*~\cite{wang2018low} significantly when the number of training samples is small, i.e. 1,2 and 5. Notably, we achieve $60.6\%$ vs $54.7\%$ state-of-the art at 1 shot, $70.3$ vs $66.8\%$ at 2 shots. This indicates that our model generates highly discriminative features by leveraging unlabeled data and word embeddings. In the challenging generalized few-shot learning setting (Figure~\ref{fig:imagenet_fsl} right), although PMN /G*~\cite{wang2018low} is quite strong by applying meta-learning~\cite{snell2017prototypical}, our model still achieves comparable results with the state-of-the-art. It is also worth noting that PMN~w/G*~\cite{wang2018low} cannot be directly applied to zero-shot learning. Hence, our approach is more versatile.

\begin{figure*}[t]
	\centering
     \includegraphics[width=0.95\linewidth]{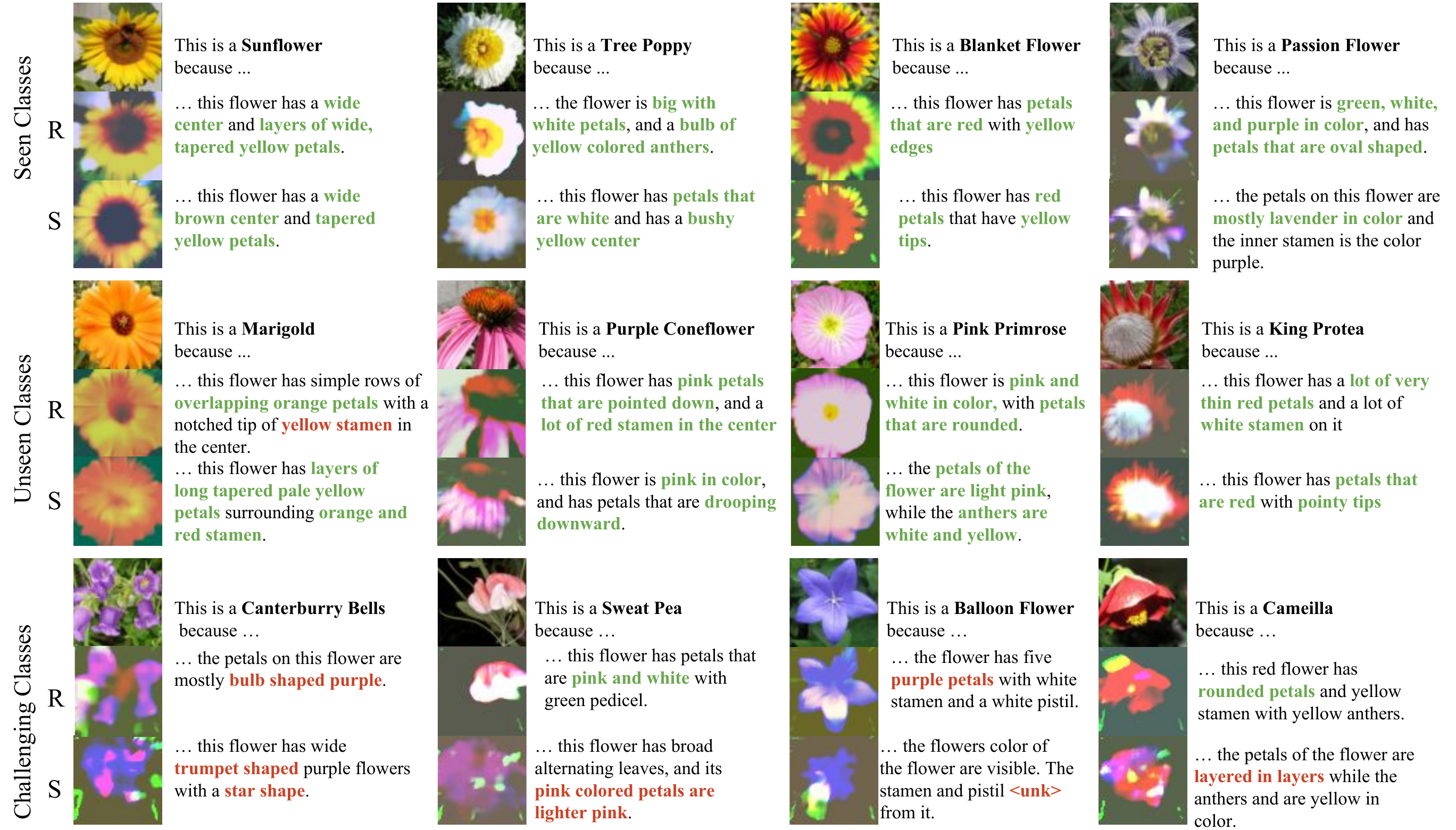}
        \caption{Interpretability: visualizations by generating images and textual explanations from real or synthetic features. For every block, the top is the target, the middle is reconstructed from the real feature (R) of the target, the bottom is reconstructed from a synthetic feature (S) from the same class. We also generate visual explanations conditioned with the predicted class and the reconstructed real or synthetic images. Top (Middle): Features come from seen (unseen) classes. Bottom: classes with a large inter-class variation lead to poorer visualizations and explanations.}
	\label{fig:inverting_features}
\end{figure*}

\subsection{Interpreting Synthesized Features}

In this section, we show that our generated features on FLO are visually discriminative and textually explainable. 

\myparagraph{Visualising generated features.}
A number of methods~\cite{dosovitskiy2016generating,mahendran2015understanding,dosovitskiy2016inverting} have explored strategies to generate images by inverting feature embeddings. We follow a strategy similar to \cite{dosovitskiy2016generating} and train a deep upconvolutional neural network to invert feature embeddings to the image pixel space. We impose a L1 loss between the ground truth image and the inverted image, as well as a perceptual loss, by passing both images through a pre-trained Resnet101, and taking an L2 loss on the feature vectors at conv5\_4 and average pooling layers. We also utilize an adversarial loss, by feeding the image and feature embedding to a discriminator, to improve our image quality. Our generator consists of a fully connected layer followed by 5 upconvolutional blocks. Each upconvolutional block contains an Upsampling layer, a 3x3 convolution, BatchNorm and ReLu non-linearity. The final size of the reconstructed image is 64x64. The discriminator processes the image through 4 downsampling blocks, the feature embedding is sent to a linear layer and spatially replicated and concatenated with the image embedding, and this final embedding is passed through a convolutional and sigmoid layer to get the probability that the sample is real or fake. We train this model on all the real feature-image pairs of the 102 classes, and use the trained generator to invert images from synthetic features. 

In Figure~\ref{fig:inverting_features}, we show generated images from real and synthetic features for comparison. 
We observe that images generated from synthetic features contain the essential attributes required for classification, such as the general color distribution and sometimes even features like the petal and stamen are visible. Also, the image quality is similar for the images generated from real and synthetic features. Interestingly, the synthetic features of unseen classes generated by our model without observing any real features from that class, i.e. ``Unseen classes'' and ``S'' row, also yield pleasing reconstructions.

As shown in ``Challenging Classes'' of Figure~\ref{fig:inverting_features}, in some cases the generated images from synthetic features lack a certain level of detail, e.g. see images for ``Balloon Flower'' and in some cases the colors do not match with the real image, e.g. see images for ``Sweat Pea''. We noticed that these correspond to classes with high inter class variation.

\myparagraph{Explaining visual features.}
We also explore generating textual explanations of our synthetic features. For this, we choose a language model \cite{hendricks2016generating}, that produces an explanation of why an image belongs to a particular class, given a feature embedding and a class label. The architecture of our model is similar to \cite{hendricks2016generating}, we use a linear layer for the feature embedding, and feed it as the start token for a LSTM. At every step in the sequence, we also feed the class embedding, to produce class relevant captions. The class embedding is obtained by training a LSTM to generate captions from images, and taking the average hidden state for images of that class. A softmax cross entropy loss is imposed on the output using the ground truth caption. Also, a discriminative loss that encourages the generated sentence to belong to the relevant class is imposed by sampling a sentence from the LSTM and sending it to a pre-trained sentence classifier. The model is trained on the dataset from \cite{RALS16}. As before, we train this model on all the real feature-caption pairs, and use it to obtain explanations for synthetic features. 

In Figure~\ref{fig:inverting_features}, we show explanations obtained from real and synthetic features.
We observe that the model generates image relevant and class specific explanations for synthetic features of both seen and unseen classes. 
For instance, a ``King Protea'' feature contains information about ``red petals and pointy tips'' while ``Purple Coneflower'' feature has information on ``pink in color and petals that are drooping downward'' which are the most visually distinguishing properties of this flower.

On the other hand, as shown at the bottom of the figure, for classes where image features lack a certain level of detail, the generated explanations have some issues such as repetitions, e.g. ``trumpet shaped'' and ``star shape'' in the same sentence and unknown words, e.g. see the explanation for ``Balloon Flower''.

\section{Conclusion}
In this work, we develop a transductive feature generating framework that synthesizes CNN image features from a class embedding. Our generated features circumvent the scarceness of the labeled training data issues and allow us to effectively train softmax classifiers. Our framework combines conditional VAE and GAN architectures to obtain a more robust generative model. We further improve VAE-GAN by adding a non-conditional discriminator that handles unlabeled data from unseen classes. The second discriminator learns the manifold of unseen classes and backpropagates the WGAN loss to feature generator such that it generalizes better to generate CNN image features for unseen classes. 

Our feature generating framework is effective across zero-shot (ZSL), generalized zero-shot (GZSL), few-shot (FSL) and generalized few-shot learning (GFSL) tasks on CUB, FLO, SUN, AWA and large-scale ImageNet datasets. Finally, we show that our generated features are visually interpretable, i.e. the generated images by by inverting features into raw image pixels achieve an impressive level of detail. They are also explainable via language, i.e. visual explanations generated using our features are class-specific.

{\small
\bibliographystyle{ieee}
\bibliography{egbib}

\begin{thebibliography}{10}\itemsep=-1pt

\bibitem{APHS13}
Z.~Akata, F.~Perronnin, Z.~Harchaoui, and C.~Schmid.
\newblock {Label embedding for attribute-based classification}.
\newblock In {\em CVPR}, 2013.

\bibitem{APHS15}
Z.~Akata, F.~Perronnin, Z.~Harchaoui, and C.~Schmid.
\newblock Label-embedding for image classification.
\newblock {\em TPAMI}, 2016.

\bibitem{arjovsky2017towards}
M.~Arjovsky and L.~Bottou.
\newblock Towards principled methods for training generative adversarial
  networks.
\newblock {\em ICLR}, 2017.

\bibitem{arjovsky2017wasserstein}
M.~Arjovsky, S.~Chintala, and L.~Bottou.
\newblock Wasserstein gan.
\newblock {\em ICML}, 2017.

\bibitem{BHJ17}
M.~Bucher, S.~Herbin, and F.~Jurie.
\newblock Generating visual representations for zero-shot classification.
\newblock {\em ICCV Workshop}, 2017.

\bibitem{CCGS16}
S.~Changpinyo, W.-L. Chao, B.~Gong, and F.~Sha.
\newblock Synthesized classifiers for zero-shot learning.
\newblock In {\em CVPR}, 2016.

\bibitem{imagenet}
J.~Deng, W.~Dong, R.~Socher, L.-J. Li, K.~Li, and L.~Fei-Fei.
\newblock {ImageNet: A Large-Scale Hierarchical Image Database}.
\newblock In {\em CVPR}, 2009.

\bibitem{dosovitskiy2016generating}
A.~Dosovitskiy and T.~Brox.
\newblock Generating images with perceptual similarity metrics based on deep
  networks.
\newblock In {\em Advances in Neural Information Processing Systems}, pages
  658--666, 2016.

\bibitem{dosovitskiy2016inverting}
A.~Dosovitskiy and T.~Brox.
\newblock Inverting visual representations with convolutional networks.
\newblock In {\em Proceedings of the IEEE Conference on Computer Vision and
  Pattern Recognition}, pages 4829--4837, 2016.

\bibitem{ESE13}
M.~Elhoseiny, B.~Saleh, and A.~Elgammal.
\newblock Write a classifier: Zero-shot learning using purely textual
  descriptions.
\newblock In {\em ICCV}, 2013.

\bibitem{FKRC18}
R.~Felix, V.~K.~B. G, I.~Reid, and G.~Carneiro.
\newblock Multi-modal cycle-consistent generalized zero-shot learning.
\newblock In {\em ECCV}, 2018.

\bibitem{finn2017model}
C.~Finn, P.~Abbeel, and S.~Levine.
\newblock Model-agnostic meta-learning for fast adaptation of deep networks.
\newblock In {\em ICML}, 2017.

\bibitem{FCSBDRM13}
A.~Frome, G.~S. Corrado, J.~Shlens, S.~Bengio, J.~Dean, M.~A. Ranzato, and
  T.~Mikolov.
\newblock Devise: A deep visual-semantic embedding model.
\newblock In {\em NIPS}, 2013.

\bibitem{FHXFG15}
Y.~Fu, T.~M. Hospedales, T.~Xiang, Z.~Fu, and S.~Gong.
\newblock Transductive multi-view zero-shot learning.
\newblock {\em TPAMI}, 37, 2015.

\bibitem{FHXG15}
Y.~Fu, T.~M. Hospedales, T.~Xiang, and S.~Gong.
\newblock Transductive multi-view zero-shot learning.
\newblock {\em TPAMI}, 2015.

\bibitem{fu2016semi}
Y.~Fu and L.~Sigal.
\newblock Semi-supervised vocabulary-informed learning.
\newblock In {\em CVPR}, 2016.

\bibitem{GPMXWDOCB14}
I.~Goodfellow, J.~Pouget-Abadie, M.~Mirza, B.~Xu, D.~Warde-Farley, S.~Ozair,
  A.~Courville, and Y.~Bengio.
\newblock Generative adversarial nets.
\newblock In {\em NIPS}, 2014.

\bibitem{gretton2007kernel}
A.~Gretton, K.~M. Borgwardt, M.~Rasch, B.~Sch{\"o}lkopf, and A.~J. Smola.
\newblock A kernel method for the two-sample-problem.
\newblock In {\em NIPS}, 2007.

\bibitem{gulrajani2017improved}
I.~Gulrajani, F.~Ahmed, M.~Arjovsky, V.~Dumoulin, and A.~Courville.
\newblock Improved training of wasserstein gans.
\newblock {\em arXiv preprint arXiv:1704.00028}, 2017.

\bibitem{HG16}
B.~Hariharan and R.~Girshick.
\newblock Low-shot visual recognition by shrinking and hallucinating features.
\newblock In {\em ICCV}, 2017.

\bibitem{hendricks2016generating}
L.~A. Hendricks, Z.~Akata, M.~Rohrbach, J.~Donahue, B.~Schiele, and T.~Darrell.
\newblock Generating visual explanations.
\newblock In {\em European Conference on Computer Vision}, pages 3--19.
  Springer, 2016.

\bibitem{jayaraman2014zero}
D.~Jayaraman and K.~Grauman.
\newblock Zero-shot recognition with unreliable attributes.
\newblock In {\em NIPS}, 2014.

\bibitem{kingma2013auto}
D.~P. Kingma and M.~Welling.
\newblock Auto-encoding variational bayes.
\newblock In {\em ICLR}, 2014.

\bibitem{kipf2016semi}
T.~N. Kipf and M.~Welling.
\newblock Semi-supervised classification with graph convolutional networks.
\newblock In {\em ICLR}, 2017.

\bibitem{koch2015siamese}
G.~Koch, R.~Zemel, and R.~Salakhutdinov.
\newblock Siamese neural networks for one-shot image recognition.
\newblock In {\em ICML Deep Learning Workshop}, 2015.

\bibitem{KXFG15}
E.~Kodirov, T.~Xiang, Z.~Fu, and S.~Gong.
\newblock Unsupervised domain adaptation for zero-shot learning.
\newblock In {\em ICCV}, 2015.

\bibitem{kodirov2017semantic}
E.~Kodirov, T.~Xiang, and S.~Gong.
\newblock Semantic autoencoder for zero-shot learning.
\newblock In {\em CVPR}, 2017.

\bibitem{Verma_2018_CVPR}
V.~Kumar~Verma, G.~Arora, A.~Mishra, and P.~Rai.
\newblock Generalized zero-shot learning via synthesized examples.
\newblock In {\em CVPR}, 2018.

\bibitem{LNH13}
C.~Lampert, H.~Nickisch, and S.~Harmeling.
\newblock Attribute-based classification for zero-shot visual object
  categorization.
\newblock {\em TPAMI}, 2013.

\bibitem{larsen2015autoencoding}
A.~B.~L. Larsen, S.~K. S{\o}nderby, H.~Larochelle, and O.~Winther.
\newblock Autoencoding beyond pixels using a learned similarity metric.
\newblock In {\em ICML}, 2016.

\bibitem{ledig2017photo}
C.~Ledig, L.~Theis, F.~Husz{\'a}r, J.~Caballero, A.~Cunningham, A.~Acosta,
  A.~P. Aitken, A.~Tejani, J.~Totz, Z.~Wang, et~al.
\newblock Photo-realistic single image super-resolution using a generative
  adversarial network.
\newblock In {\em CVPR}, 2017.

\bibitem{lei2015predicting}
J.~Lei~Ba, K.~Swersky, S.~Fidler, et~al.
\newblock Predicting deep zero-shot convolutional neural networks using textual
  descriptions.
\newblock In {\em ICCV}, 2015.

\bibitem{li2015generative}
Y.~Li, K.~Swersky, and R.~Zemel.
\newblock Generative moment matching networks.
\newblock In {\em ICML}, 2015.

\bibitem{mahendran2015understanding}
A.~Mahendran and A.~Vedaldi.
\newblock Understanding deep image representations by inverting them.
\newblock In {\em Proceedings of the IEEE conference on computer vision and
  pattern recognition}, pages 5188--5196, 2015.

\bibitem{MSCCD13}
T.~Mikolov, I.~Sutskever, K.~Chen, G.~S. Corrado, and J.~Dean.
\newblock Distributed representations of words and phrases and their
  compositionality.
\newblock In {\em NIPS}, 2013.

\bibitem{conditionalgans}
M.~Mirza and S.~Osindero.
\newblock Conditional generative adversarial nets.
\newblock {\em arXiv preprint arXiv:1411.1784}, 2014.

\bibitem{miyato2018spectral}
T.~Miyato, T.~Kataoka, M.~Koyama, and Y.~Yoshida.
\newblock Spectral normalization for generative adversarial networks.
\newblock In {\em ICLR}, 2018.

\bibitem{mukherjee2016gaussian}
T.~Mukherjee and T.~Hospedales.
\newblock Gaussian visual-linguistic embedding for zero-shot recognition.
\newblock In {\em EMNLP}, 2016.

\bibitem{OxfordFlowersDataset}
M.-E. Nilsback and A.~Zisserman.
\newblock Automated flower classification over a large number of classes.
\newblock In {\em ICCVGI}, 2008.

\bibitem{NMBSSFCD14}
M.~Norouzi, T.~Mikolov, S.~Bengio, Y.~Singer, J.~Shlens, A.~Frome, G.~Corrado,
  and J.~Dean.
\newblock Zero-shot learning by convex combination of semantic embeddings.
\newblock In {\em ICLR}, 2014.

\bibitem{PH12}
G.~Patterson and J.~Hays.
\newblock Sun attribute database: Discovering, annotating, and recognizing
  scene attributes.
\newblock In {\em CVPR}, 2012.

\bibitem{qi2018low}
H.~Qi, M.~Brown, and D.~G. Lowe.
\newblock Low-shot learning with imprinted weights.
\newblock In {\em CVPR}, 2018.

\bibitem{qiao2018few}
S.~Qiao, C.~Liu, W.~Shen, and A.~L. Yuille.
\newblock Few-shot image recognition by predicting parameters from activations.
\newblock In {\em CVPR}, 2018.

\bibitem{RMC16}
A.~Radford, L.~Metz, and S.~Chintala.
\newblock Unsupervised representation learning with deep convolutional
  generative adversarial networks.
\newblock In {\em ICLR}, 2016.

\bibitem{ravi2016optimization}
S.~Ravi and H.~Larochelle.
\newblock Optimization as a model for few-shot learning.
\newblock In {\em ICLR}, 2016.

\bibitem{RALS16}
S.~Reed, Z.~Akata, H.~Lee, and B.~Schiele.
\newblock Learning deep representations of fine-grained visual descriptions.
\newblock In {\em CVPR}, 2016.

\bibitem{RAYLSL16}
S.~Reed, Z.~Akata, X.~Yan, L.~Logeswaran, B.~Schiele, and H.~Lee.
\newblock Generative adversarial text to image synthesis.
\newblock In {\em ICML}, 2016.

\bibitem{MES13}
M.~Rohrbach, S.~Ebert, and B.~Schiele.
\newblock Transfer learning in a transductive setting.
\newblock In {\em NIPS}, 2013.

\bibitem{RT15}
B.~Romera-Paredes and P.~H. Torr.
\newblock An embarrassingly simple approach to zero-shot learning.
\newblock {\em ICML}, 2015.

\bibitem{snell2017prototypical}
J.~Snell, K.~Swersky, and R.~Zemel.
\newblock Prototypical networks for few-shot learning.
\newblock In {\em NIPS}, 2017.

\bibitem{song2018transductive}
J.~Song, C.~Shen, Y.~Yang, Y.~Liu, and M.~Song.
\newblock Transductive unbiased embedding for zero-shot learning.
\newblock In {\em CVPR}, 2018.

\bibitem{verma2017simple}
V.~K. Verma and P.~Rai.
\newblock A simple exponential family framework for zero-shot learning.
\newblock In {\em ECML}, 2017.

\bibitem{vinyals2016matching}
O.~Vinyals, C.~Blundell, T.~Lillicrap, D.~Wierstra, et~al.
\newblock Matching networks for one shot learning.
\newblock In {\em NIPS}, 2016.

\bibitem{Wang_2018_CVPR}
X.~Wang, Y.~Ye, and A.~Gupta.
\newblock Zero-shot recognition via semantic embeddings and knowledge graphs.
\newblock In {\em CVPR}, 2018.

\bibitem{wang2018low}
Y.~Wang, R.~Girshick, M.~Hebert, and B.~Hariharan.
\newblock Low-shot learning from imaginary data.
\newblock In {\em CVPR}, 2018.

\bibitem{CaltechUCSDBirdsDataset}
P.~Welinder, S.~Branson, T.~Mita, C.~Wah, F.~Schroff, S.~Belongie, and
  P.~Perona.
\newblock {Caltech-UCSD Birds 200}.
\newblock Technical Report CNS-TR-2010-001, Caltech, 2010.

\bibitem{xian2018zero}
Y.~Xian, C.~H. Lampert, B.~Schiele, and Z.~Akata.
\newblock Zero-shot learning-a comprehensive evaluation of the good, the bad
  and the ugly.
\newblock {\em TPAMI}, 2018.

\bibitem{XLSA18}
Y.~Xian, T.~Lorenz, B.~Schiele, and Z.~Akata.
\newblock Feature generating networks for zero-shot learning.
\newblock In {\em CVPR}, 2018.

\bibitem{ye2017zero}
M.~Ye and Y.~Guo.
\newblock Zero-shot classification with discriminative semantic representation
  learning.
\newblock In {\em CVPR}, 2017.

\bibitem{zhang2016learning}
L.~Zhang, T.~Xiang, and S.~Gong.
\newblock Learning a deep embedding model for zero-shot learning.
\newblock In {\em CVPR}, 2017.

\bibitem{ZV15}
Z.~Zhang and V.~Saligrama.
\newblock Zero-shot learning via semantic similarity embedding.
\newblock In {\em ICCV}, 2015.

\bibitem{CycleGAN2017}
J.-Y. Zhu, T.~Park, P.~Isola, and A.~A. Efros.
\newblock Unpaired image-to-image translation using cycle-consistent
  adversarial networks.
\newblock In {\em ICCV}, 2017.

\bibitem{Zhu_2018_CVPR}
Y.~Zhu, M.~Elhoseiny, B.~Liu, X.~Peng, and A.~Elgammal.
\newblock A generative adversarial approach for zero-shot learning from noisy
  texts.
\newblock In {\em CVPR}, 2018.

\end{thebibliography}
}

\end{document}